\RequirePackage{fix-cm}
\documentclass[smallcondensed, natbib]{svjour3}     
\smartqed  
\usepackage[table,xcdraw]{xcolor}
\usepackage{graphicx}
\usepackage{amsfonts}
\usepackage{subfig}
\usepackage{mathptmx}      
\usepackage{latexsym}
\journalname{}
\begin{document}

\title{A Survey on Deep Learning Techniques for Video Anomaly Detection}

\author{Jessie James P. Suarez         \and
        Prospero C. Naval, Jr. 
}


\institute{Jessie James P. Suarez \at
              Computer Vision and Machine Intelligence Group \\
              University of the Philippines, Diliman \\
              \email{jpsuarez@up.edu.ph}           
           \and
           Prospero C. Naval, Jr \at
              Computer Vision and Machine Intelligence Group \\
              University of the Philippines, Diliman \\
              \email{pcnaval@dcs.upd.edu.ph}  
}

\date{Received: date / Accepted: date}

\maketitle

\begin{abstract}
Anomaly detection in videos is a problem that has been studied for more than a decade. This area has piqued the interest of researchers due to its wide applicability. Because of this, there has been a wide array of approaches that have been proposed throughout the years and these approaches range from statistical-based approaches to machine learning-based approaches. Numerous surveys have already been conducted on this area but this paper focuses on providing an overview on the recent advances in the field of anomaly detection using Deep Learning. Deep Learning has been applied successfully in many fields of artificial intelligence such as computer vision, natural language processing and more. This survey, however, focuses on how Deep Learning has improved and provided more insights to the area of video anomaly detection. This paper provides a categorization of the different Deep Learning approaches with respect to their objectives. Additionally, it also discusses the commonly used datasets along with the common evaluation metrics. Afterwards, a discussion synthesizing all of the recent approaches is made to provide direction and possible areas for future research.
\end{abstract}

\section{Introduction}
\begin{figure}[h]
  \centering
  \includegraphics[scale=0.3]{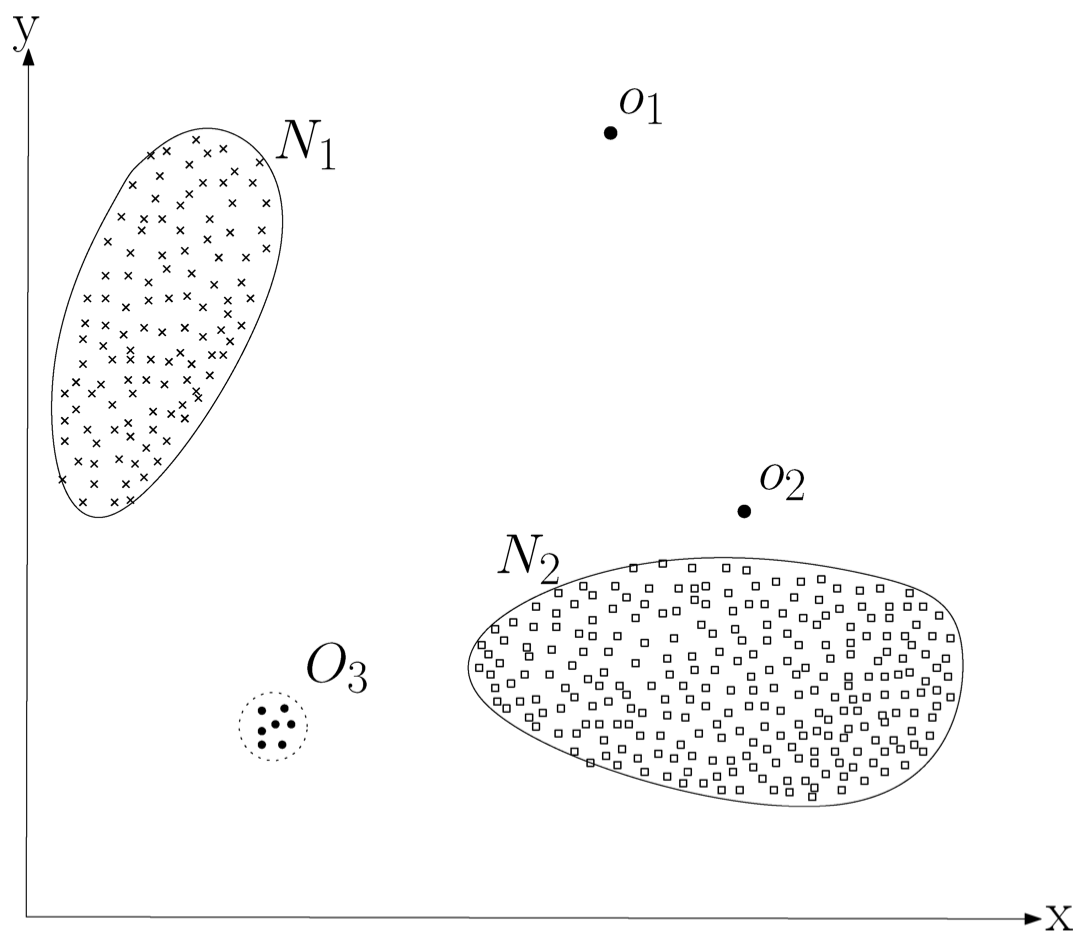}
  \caption{Simple Anomaly Case by Chandola et. al. \citeyear{chandola_2019_anomaly_detection}}
  \label{pic_anomly_simple}
\end{figure}

Surveillance videos have been increasingly present in various establishments in order to monitor human activity and prevent crime from happening. It goes without saying that there needs to be someone behind watching the videos and signaling an alert whenever something different from normal is happening. However, these events do not happen very often and that most of the time, the person monitoring these videos would see nothing out of the ordinary \citep{sultani_2018_real_world_anomaly}. These unusual events can be thought of as \textit{anomalies} which can be defined as patterns that do not conform to what is considered normal. The task of finding these nonconforming patterns is called \textit{anomaly detection} \citep{chandola_2019_anomaly_detection}. Because of this, researchers have been trying to create a robust anomaly detection algorithms that can automate the process of monitoring and detection of unusual events in surveillance videos. An example of a simple anomaly case can be seen in Fig. \ref{pic_anomly_simple} where the normal regions are denoted by $N$ and anomalies are those denoted by $O$. As seen in the figure, anomalies tend to clearly lie outside what is normal. However, these anomalies can, in fact, be close to normality which is illustrated by $O_2$

Anomaly detection is a challenging task due to number of reasons: first, the definition of an anomaly may vary from one context to another \citep{medel_2016_predictive_cnn_lstm, sabokrou_2017_deep_cascade}. Second, the different possibilities of what constitute an anomaly might be are boundless \citep{luo_2019_sparse_coding_dnn}. Third, anomalous data points, especially with real-world data, tend to lie closely to what might be defined as normal \citep{vu_2017_energy_based_models}. Lastly, extracting robust features from the data even if anomalies seldom appear \citep{ribeiro_2018_deep_ae}. The mentioned list does not entirely capture all of the possible reasons which make the problem hard but these main points are what researchers have been considering for the past years when proposing new solutions to the problem. 

Around a decade ago, most of the researchers have focused on trajectory-based anomaly detection \citep{fan_2011_spatiotemporal_context, calderara_2011_spectral_graph, tung_2010_goal_based_trajectory, li_2013_trajectory_sparse_reconstruction}. The main idea is if the objects of interest are not following the learned normal trajectories, the video will be tagged as an anomaly. However, one major drawback of this approach is occlusion since the approach heavily relies on continuously monitoring the objects of interest \citep{sabokrou_2017_deep_cascade, narasimhan_2018_sparse_denoising_ae}. Due to these drawbacks, there was an emphasis on using low-level features for feature extraction instead \citep{sabokrou_2017_deep_cascade}. These approaches based on low-level features rely on the use of appearance, motion, and texture features \citep{mehran_2009_social_force, li_2014_detection_in_crowded_scenes, zhang_2016_combining_appearance_and_motion, wang_2018_local_motion_ocelm, kim_2009_spacetime_mrf, benezeth_2009_spatiotemporal_cooccurrences}. Various representations have been used in order to represent these aspects of the video such as in the approach of \cite{mehran_2009_social_force} where they used social force maps to model motion of the crowds. Similarly, pixel-motion properties were used by \cite{benezeth_2009_spatiotemporal_cooccurrences} to model behavior. Meanwhile, \cite{kim_2009_spacetime_mrf} made use of optical flows which are then used as inputs to the mixture of probabilistic principal component analysis (MPPCA) model, thus, creating a more compact feature representation. However, features based on motion are not enough which is why there were proposed approaches that make use of both. An example is the approach of \cite{li_2014_detection_in_crowded_scenes} where their approach makes use of mixture of dynamic textures (MDTs) that utilize temporal normalcy and discriminant saliency detectors to model spatial normalcy. Likewise, \cite{zhang_2016_combining_appearance_and_motion} used support vector data description for spatial features and optical flow for motion features. In contrast, \cite{wang_2018_local_motion_ocelm} used spatially localized histogram of optical flows and uniform local gradient pattern-based optical flows. Most of these techniques and methods, specifically on these "traditional" approaches, have been discussed in great detail in the works of \cite{kaur_2018_overview_of_anomaly_detection, li_2016_anomaly_detection_techniques, popola_2012_abnormal_review}.

Despite the proven success of these traditional approaches on benchmark datasets, they are still ineffective when used in a different domain. Furthermore, they are unable to adapt to anomalies that they have never seen before \citep{hu_2016_deep_incremental_sfa, medel_2016_predictive_cnn_lstm}. For these reasons, recent works have mostly explored the use of Deep Neural Networks for the task of anomaly detection. These neural networks automatically learn useful and discriminant features on their own which removes the hassle of creating handcrafting features \citep{krizhevsky_2012_imagenet_classification}. This also makes it more adaptive when used on different domains. Deep learning was proven to be effective for a variety of computer vision tasks such as feature extraction in images \citep{yan_2016_deep_cnn_feature}, image classification \citep{krizhevsky_2012_imagenet_classification}, object detection \citep{zoph_2018_transferrable_architectures}, video analysis \citep{mei_2017_deep_learning_for_video}, face detection \citep{lopes_2017_cnn_face}, visual question answering \citep{malinowski_2017_vqa} and many other tasks.

As mentioned previously, there are existing works that have discussed various anomaly detection methods for videos \citep{kaur_2018_overview_of_anomaly_detection, li_2016_anomaly_detection_techniques, popola_2012_abnormal_review}. However, due to the recent traction in the use of deep learning techniques on this field, the goal of this paper is to provide a closer look into these deep learning techniques. This entails providing organization as to how the approaches are related to one another, the rationale as to why these methods have been proposed, and summarizing the conclusions which they have presented in a clear manner. In addition, it would also be necessary to discuss datasets and evaluation metrics which have mostly been used by these approaches. It would also be insightful to determine how these datasets and metrics would scale well when dealing with real-world anomaly detection. Different researchers have created different environmental setups  making some of them incomparable. Thus, the performances of the approaches discussed will not be included to avoid confusion and misinterpretation.

The paper is organized as follows: the first section serves as an introduction to the survey. Second, deep learning anomaly detection techniques will be discussed in detail. Third, the mostly used datasets will be tackled. Fourth, the commonly used evaluation metrics will be presented. Fifth, a section for discussion is allocated to synthesize all of the approaches and datasets mentioned. Lastly, the concluding remarks coupled with recommendations as to what directions this area of research could possibly go.

\section{Deep Learning in Anomaly Detection for Videos}
Deep learning techniques mostly focus on creating new architectures or crafting components that can be suitable for a specific problem. Since deep learning methods have been successful in a number of varied use cases \citep{yan_2016_deep_cnn_feature, krizhevsky_2012_imagenet_classification, zoph_2018_transferrable_architectures, mei_2017_deep_learning_for_video, lopes_2017_cnn_face, malinowski_2017_vqa}, most of these networks or architectures might be similar to each other. An example of which would be with \cite{krizhevsky_2012_imagenet_classification} where they used Convolutional Neural Networks for image classification. However, almost the same network is also used for face recognition \citep{lopes_2017_cnn_face}. Because of this, the presented categories below would group these approaches specifically with respect to their final objectives instead of network architecture or learning strategy. Examples of these include using reconstruction error or providing an anomaly score. In line with this, there are four (4) identified categories namely: using reconstruction error or reconstruction-based methods, framing the problem as a classification problem, predicting future frames, and computing for an anomaly score. A quick summary of all these techniques are provided in Table \ref{deep_learning_techniques}.

\subsection{Using Reconstruction Error} \label{subsection_reconstruction}
Reconstruction error has already been used in various traditional anomaly detection techniques \citep{popola_2012_abnormal_review}. The basic assumption of using reconstruction error is that the reconstruction error for normal samples would be lower since they are closer to the training data. On the other hand, the reconstruction error is assumed or expected to be higher for samples which are not normal \citep{gong_2019_memory_deep_ae, sabokrou_2016_sparsity_reconstruction_error_ae}. 

More formally, let $x$ be a video segment or video frame and let $g$ be a neural network that reconstructs $x$. The reconstruction error can be defined as a function $f$ such that is computes for error between $x$ (the original input), and $g(x)$ which is the reconstruction Eqn \ref{eq_reconstruction}. This concept has been extended recently by making use of deep learning techniques to reconstruct various scenes.

\begin{equation} \label{eq_reconstruction}
    e = f(x, g(x))
\end{equation}

\begin{figure}[h]
  \centering
  \includegraphics[scale=0.4]{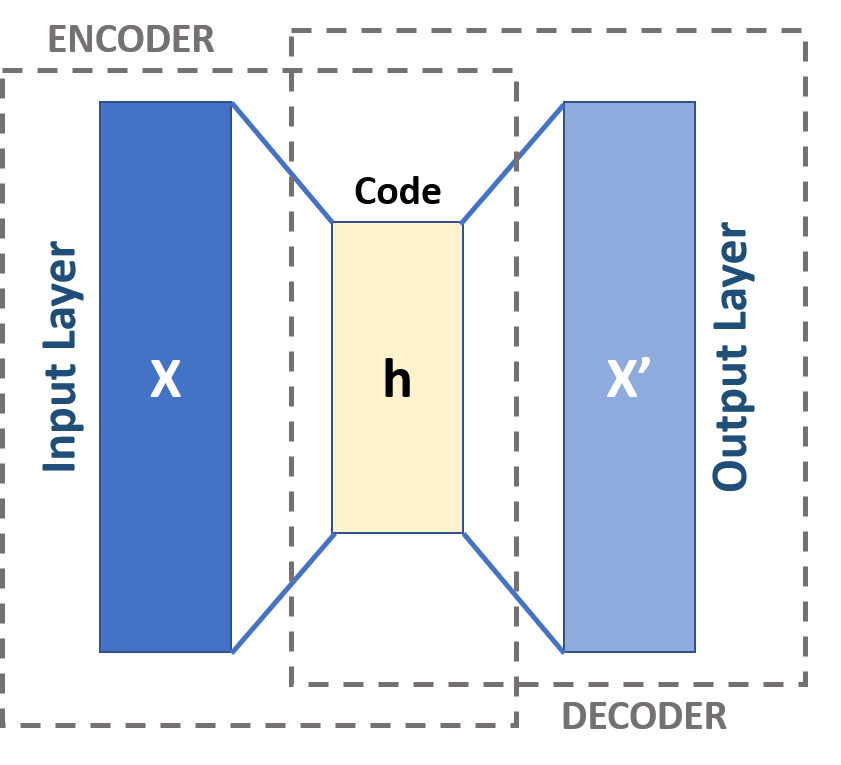}
  \caption{Autoencoder Diagram by Michaela \cite{massi_2019_pic_ae} via Wikimedia Commons}
  \label{pic_autoencoder}
\end{figure}

Different from usual feedforward networks, one type of neural network that is able to reconstruct input data is called an autoencoder. The autoencoder is a neural network that has the capability to encode an input into a more compact representation while retaining important and discriminative features. It also has the ability to decode this particular encoding back to its original form \citep{baldi_2011_autoencoders}. A visual schematic of an autoencoder is shown in Fig. \ref{pic_autoencoder} where the diagram illustrates a simple architecture of an encoder where the left-hand side is the input to the autoencoder $X$, the middle portion is the encoded representation (sometimes called the latent vector or code) of $X$, and the right-hand side is the decoded encoding called $X'$.

Most approaches whose goal of using reconstruction error as a means to identify anomalies base their method on autoencoders. One such method is introduced by \cite{hasan_2016_temporal_regularity} where they posited that in comparison to sparse coding, the objective function of an autoencoder is more efficient. They have also said that it is able to preserve spatio-temporal information while encoding dynamics. Their approach made use of combining 2D convolutions to autoencoders wherein the 2D convolutions take as input specific raw video segments. Conventionally, inputs to a Convolutional Neural Network is a 2D image having the third channel as the color channel \citep{krizhevsky_2012_imagenet_classification}. However, in their approach, the third dimension is instead composed of stacked grayscale frames, allowing the model to encode both spatial and some temporal information for reconstruction. 

Similarly, the work of \cite{medel_2016_predictive_cnn_lstm} also framed the problem as a reconstruction problem. The approach makes use of a convolutional long short-term memory wherein the Long Short-Term Memory (LSTM) Network is a type of neural network that is capable of learning long-term dependencies of the data \citep{hochreiter_1997_lstm}. Despite not being explicitly an autoencoder, their approach also makes use of an encoder-decoder sturture. Given an input sequence of video frames, the convolutional long short-term memory extracts relevant features along the spatial and temporal dimension in such a way that the last time step is used as the encoding. The decoder unravels the encoding and then reconstructs the frames which can then be used to compute the reconstruction error for anomaly detection.

The proposed approach of \cite{ribeiro_2018_deep_ae} closely resembles that of \cite{hasan_2016_temporal_regularity}. The main difference is that the low-level features such as optical flow and edges are used as inputs alongside the raw frames. In addition, they have also presented how these features affect the convolutional autoencoder with regard to detecting anomalies.

Another method was proposed by \cite{sabokrou_2016_sparsity_reconstruction_error_ae} where they have used two different autoencoders for the task: one is a regular autoencoder and the other is a sparse autoencoder. A sparse autoencoder is an autoencoder but has an additional sparsity penalty. This penalty encourages fewer neurons to activate. This constraint allows the network to learn relevant information without reducing the number of nodes in the hidden layers. Their approach involves two steps, the first step is to compute the sparsity value from cubic patches of the videos, if it is below a specific sparsity threshold, another set of patches are extracted around that patch for reconstruction. 

According to \cite{zhao_2017_spatiotemporal_ae}, the approach of \cite{hasan_2016_temporal_regularity} which makes use of temporal cuboids by stacking frames in the third dimension, does not necessarily retain the temporal information. Based on their work, a reason for this is that 2D convolutions operate on the frames spatially. Putting this in the perspective of the approach of \cite{hasan_2016_temporal_regularity}, the third channel is represented along each of the channels of the first feature map which rarely preserves temporal information. To solve this, \cite{zhao_2017_spatiotemporal_ae} proposed the use of 3D convolutions as a means to retain temporal information during the convolution process. Since it is data intensive, they have also applied data augmentation to increase their samples.

As claimed in the work of  \cite{zhou_2019_anomalynet}, one weakness of the approach of  \cite{medel_2016_predictive_cnn_lstm} is that spatial and temporal aspects of the inputs are encoded separately by the convolutions and the long short-term memory. This implies a broken relationship between the two during the encoding process. Furthermore, it was also stated by \cite{zhou_2019_anomalynet} that the approach proposed by \cite{medel_2016_predictive_cnn_lstm} was not able to make use of existing pre-trained networks. These networks have shown remarkably improved performances once it has been applied to other domains. Hence, their proposed method makes use of a feature learning subnetwork that combines motion and appearance features into an image. Afterwards, it is then used as an input to a pretrained network for feature extraction. Moreover, they have proposed a novel subnetwork called sparse coding to network (SC2Net) to compute for the sparsity loss and reconstruction loss from the extracted features.

Among all of the approaches, \cite{gong_2019_memory_deep_ae} have posited that most of the works on reconstruction generally assume that the anomalous instances will have a high reconstruction error. Based on these works, this assumption does not necessarily hold true mainly because there might be instances where an autoencoder is able to generalize well. This poses a problem since it might accurately reconstruct anomalous instances as well. To mitigate this problem, they have introduced a new autoencoder which has the capability to store encodings into memory. The main difference from previous approaches is that instead of directly feeding the encoding to the decoder, the encoding is treated as a query. This query is expected to return closest normal patterns in memory which is instead used for decoding. In the event that an anomaly is to be reconstructed, it would have a high reconstruction error because the memory only has normal memory items.

\subsection{Using Future Frame Prediction}

A different perspective on the problem was presented by \cite{liu_2018_future_frame}. They support the claim of  \cite{gong_2019_memory_deep_ae} stating that autoencoders might also accurately reconstruct anomalous frames. Since anomalies can be viewed as events that do not conform with certain expectations, \cite{liu_2018_future_frame} suggested a frame prediction approach might be a more natural way to view the problem. Mathematically speaking, given $x_t$ which is the video segment or frame $x$ at time $t$, future frame prediction can be expressed as a function $h$ predicting the next segment as shown in Eqn \ref{eq_future}.

\begin{equation}\label{eq_future}
    x_{t+1} = h(x_t)
\end{equation}

In deep learning, there is a specific type of neural network is used for generating new data with the same statistics as the training data. This network which is called generative adversarial network (GAN) \citep{goodfellow_2014_gan}. This architecture has two main (2) parts. The first one is a generator whose job is to mimic the original data distribution. Meanwhile, the second network is called a discriminator that gives a probability of whether or not the input is coming from the generator. 

The approach of \cite{liu_2018_future_frame} made use of a generator-discriminator structure, likened to that of a generative adversarial network. They used the U-Net architecture \citep{ronneberger_2015_unet} for future frame prediction as the generator because of its exemplary performance in image-to-image translation. While the discriminator at the end of the network determines whether or not the predicted frame is anomalous. 

Some works on reconstruction also have the capability for predicting future frames such as in the work of \cite{hasan_2016_temporal_regularity}. Their approach has the ability to encode both spatial and temporal aspects of the video by allowing the autoencoder to learn it from a sequence of video segments (discussed in more detail in Section \ref{subsection_reconstruction}). It is because of this exact same reason that it can also predict future and past frames given a center frame. Based on their methodology, by padding the center frame with zero values, their model can extrapolate the near future and near past of the center frame. 

Moreover, some of the previous works actually leverage future frame prediction in the process of reconstructing the current frame. An example of this is the work of  \cite{zhao_2017_spatiotemporal_ae} where their network learns the future frames along with the task of reconstruction in a different branch of the network. Similarly, \cite{medel_2016_predictive_cnn_lstm} also has a separate branch in parallel that learns how to predict the future. Despite their similarities, they both have big differences as to how future frame prediction is used. \cite{medel_2016_predictive_cnn_lstm} makes use of future frame prediction to identify interest points within the video. On the contrary, in the approach of  \cite{zhao_2017_spatiotemporal_ae}, the future frame is actually included in the computation of the loss to guide the network to extract temporal features. In addition, it is also included in the reconstruction score which combines the prediction loss and the reconstruction loss.

\subsection{Using Classifiers}

Despite the sophisticated methods that rely mainly on reconstruction loss and future frame prediction, there are also still a handful of approaches that cast the problem as a classification problem. The classification problem can be viewed as a function $j$ that takes as its input a frame or video segment $x$ whose output $y$ is a class or category as seen in Eqn \ref{eq_classification}.

\begin{equation}\label{eq_classification}
    y = j(x), y \in \mathbb{R}
\end{equation}

Because of imbalanced datasets, these methods focus mostly on how to create compact, efficient, and robust features. The approach of \cite{sabokrou_2017_deep_cascade} tries to solve this problem by proposing a competitive cascade of deep neural networks. The cascade is composed of two stages where the first stage is a small stack of autoencoders which hierarchically models the normality of the video patches. The other one is a Convolutional Neural Network which takes as input video patches that the autoencoders could not handle and would need further probing. The classifier used for the approach is a Gaussian Classifier.

\begin{table}[]
\small
\centering
\hspace*{-0.5cm}
\caption{Summary of Methods and Contributions}
\label{deep_learning_techniques}
\hspace*{-0.3cm}
\begin{tabular}{|l|l|l|l|}
\hline
\rowcolor[HTML]{000000} 
\multicolumn{1}{|c|}{\cellcolor[HTML]{000000}{\color[HTML]{FFFFFF} \textbf{Year}}} & \multicolumn{1}{c|}{\cellcolor[HTML]{000000}{\color[HTML]{FFFFFF} \textbf{Author}}} & \multicolumn{1}{c|}{\cellcolor[HTML]{000000}{\color[HTML]{FFFFFF} \textbf{Type}}} & \multicolumn{1}{c|}{\cellcolor[HTML]{000000}{\color[HTML]{FFFFFF} \textbf{Main Contribution}}} \\ \hline
2016 & Medel et. al & \begin{tabular}[c]{@{}l@{}}Reconstruction\\ \& Future Frame\end{tabular} & \begin{tabular}[c]{@{}l@{}}Convolutional Long \\ Short-Term Memory\end{tabular} \\ \hline
2016 & Hasan et. al. & Reconstruction & \begin{tabular}[c]{@{}l@{}}Fully 2D Convolutional\\ Autoencoder\end{tabular} \\ \hline
2016 & Sabokrou et. al. & Reconstruction & \begin{tabular}[c]{@{}l@{}}Sparse Autoencoder +\\ Autoencoder\end{tabular} \\ \hline
2016 & Hu et. al. & Scoring & \begin{tabular}[c]{@{}l@{}}Deep Neural Network +\\ Slow Feature Analysis\end{tabular} \\ \hline
2017 & Narasimhan et. al. & Classification & \begin{tabular}[c]{@{}l@{}}Sparse Denoising \\ Autoencoders\end{tabular} \\ \hline
2017 & Sabokrou et. al. & Classification & \begin{tabular}[c]{@{}l@{}}Cascade of Deep \\ Convolutional Neural \\ Networks + \\ Autoencoders\end{tabular} \\ \hline
2017 & Zhao et. al. & \begin{tabular}[c]{@{}l@{}}Reconstruction\\ \& Future Frame\end{tabular} & \begin{tabular}[c]{@{}l@{}}Spatiotemporal\\ Autoencoder\end{tabular} \\ \hline
2018 & Sabokrou et. al. & Classification & Deep-Anomaly \\ \hline
2018 & Sultani et. al. & Scoring & \begin{tabular}[c]{@{}l@{}}Multiple-Instance \\ Learning\end{tabular} \\ \hline
2018 & Ribeiro et. al. & Reconstruction & \begin{tabular}[c]{@{}l@{}}Low-level Features + \\ 2D Convolutional \\ Autoencoder\end{tabular} \\ \hline
2018 & Liu et. al. & Future Frame & \begin{tabular}[c]{@{}l@{}}Future Frame using \\ U-Net\end{tabular} \\ \hline
2019 & Landi et. al. & Scoring & \begin{tabular}[c]{@{}l@{}}Localization before\\ Feature Extraction\end{tabular} \\ \hline
2019 & Sabzailan et. al. & Scoring & \begin{tabular}[c]{@{}l@{}}Traditional + Deep\\ Learning Features\end{tabular} \\ \hline
2019 & Zhu et. al. & Scoring & \begin{tabular}[c]{@{}l@{}}Optical Flow as inputs\\ to Multiple-Instance\\ Learning\end{tabular} \\ \hline
2019 & Zhou et. al. & Reconstruction & \begin{tabular}[c]{@{}l@{}}AnomalyNet: a unified\\ approach\end{tabular} \\ \hline
2019 & Gong et. al. & Reconstruction & \begin{tabular}[c]{@{}l@{}}Autoencoder + memory\\ module + attention-based \\ addressing\end{tabular} \\ \hline
2019 & Lin. et. al. & Scoring & \begin{tabular}[c]{@{}l@{}}Multiple-Instance \\ Learning + Social Force \\ Maps\end{tabular} \\ \hline
2019 & Santos et. al. & Classification & \begin{tabular}[c]{@{}l@{}}Transfer Learning + \\ Transfer Component\\ Analysis\end{tabular} \\ \hline
2019 & Luo et. al. & Scoring & \begin{tabular}[c]{@{}l@{}}Sparse Coding-inspired\\ Deep Neural Network\end{tabular} \\ \hline
2019 & Ionescu et. al. & Classification & \begin{tabular}[c]{@{}l@{}}Object-Centric \\ Convolutional \\ Autoencoders\end{tabular} \\ \hline
2019 & Xu et. al. & Classification & \begin{tabular}[c]{@{}l@{}}Adaptive Intra-Frame\\ Classification Network\end{tabular} \\ \hline
2020 & Fan et. al. & Scoring & \begin{tabular}[c]{@{}l@{}}Gaussian Mixture\\ Fully Convolutional\\ Variational Autoencoders\end{tabular} \\ \hline
\end{tabular}
\end{table}

On the other hand, \cite{narasimhan_2018_sparse_denoising_ae} proposed a method that makes use of local and global descriptors whose aim is utilize both spatial and temporal domains. For local features, they made use of an image similarity metric on the video cubic patches to represent the temporal and spatial features. Meanwhile, the global features are represented by the latent vector of the trained autoencoders. After creating both local and global features, it is then fed to an autoencoder which selects important features that are discriminative enough for anomaly detection. Finally, these features are fed into Gaussian classifiers separately for local and global descriptors and then combined to detect anomalies.

Most of the above mentioned methods, even those in the previous sections, make use of Convolutional Neural Networks. However, \cite{sabokrou_2018_deep_anomaly} has mentioned problems with regard to using these networks, one of which is that these networks are too inefficient for patch-based methods. Examples of approaches that made use of patches are as follows: \cite{narasimhan_2018_sparse_denoising_ae, sabokrou_2016_sparsity_reconstruction_error_ae, sabzalian_2019_deep_and_sparse_features, sabokrou_2018_deep_anomaly, medel_2016_predictive_cnn_lstm}. For this reason, they have proposed a possible solution to the problem which makes use of the discriminative power of a pre-trained model without having to tweak it \cite{sabokrou_2018_deep_anomaly}. More specifically, they use the intermediate layer to generate the features that will be fed to a Gaussian Classifier. In the event that a low confidence is generated by the classifier, it is sent to another convolutional layer on top of the best intermediate layer for further probing. 

Similar to \cite{sabokrou_2018_deep_anomaly}, the proposed approach of \cite{dossantos_2019_generalization_of_feature_embeddings} took advantage of the available pre-trained models. They have investigated the generalization of feature spaces of Convolutional Neural Networks without requiring additional labels. In their experiments, they used transfer component analysis \citep{pan_2011_transfer_component_analysis} which attempts to learn a certain subspace that is shared by different domains. They have concluded that generalization through different domains.

Most of the methods mentioned previously make use of extracting either global or local features without taking the objects of interest into account. The approach of \cite{ionescu_2019_object_centric} makes use of a single-shot detector (SSD) \citep{liu_2016_ssd} on each frame of the video. After isolating the objects, a convolutional autoencoder is used to learn deep unsupervised features thereby allowing the algorithm to focus on the objects in the scene. Furthermore, they have instead casted the problem of anomaly detection as a multi-class classification problem rather than an unbalanced binary classification problem or a one-class problem. To generate the artificial classes, they have used clustering on the set of features generated by the convolutional autoencoder where each cluster represents a different type of normality. A one-versus-rest classifier is trained which discriminates between the clusters. If the highest classification score is negative, meaning the sample does not belong to any cluster, it is tagged as anomalous. 

Similar to \cite{ionescu_2019_object_centric}, \cite{xu_2020_adaptive_intraframe} also framed the problem as a multi-class classification problem as opposed to either a one-class or a binary classification problem. In line with this, they also took note of the fact that most of the previous approaches were able to effectively identify subregions representations of anomalies. However, for most of the approaches, there is a wide array of inputs and outputs such as optical flows, patches, or gradients. This inspired the approach of \cite{xu_2020_adaptive_intraframe} which tries to unify all of these approach by creating a network called the adaptive intraframe classification network that takes the raw inputs, computes for motion and appearance features, and determines whether or not the sample is anomalous. 

\subsection{Using Scoring Methods}

Some researchers have instead, framed the problem as a regression problem wherein the goal is to provide an anomaly score which will then be used as a means to determine whether or not a video segment or a frame is anomalous \citep{landi_2019_anomaly_locality, sultani_2018_real_world_anomaly}. The scoring methods can be viewed as a function $k$ such that it takes a video segment or frame $x$ as its input. It outputs a real number $z$ representing the anomaly score as seen in Eqn \ref{eq_scoring}.

\begin{equation}\label{eq_scoring}
    z = k(x), z \in \mathbb{R}
\end{equation}

The proposed approach of \cite{hu_2016_deep_incremental_sfa}, makes use of their novel sum squared derivative to score the features generated by their approach. This basically determines if the sequence of frames is anomalous. Prior to their scoring method, they combined both deep learning and slow feature analysis \citep{wiskott_2002_slow_feature_analysis} in order to learn semantic-level representations given raw video frames. It is also worth noting that their approach has an online variant, thereby making their approach adaptive.

The approach of \cite{sultani_2018_real_world_anomaly} made use of a multiple instance learning to identify anomalies in video segments based on weakly-labelled videos (labels are on a video-level and not frame-level). Their approach uses C3D, a 3D Convolutional Neural Network that learns spatiotemporal features by exposing the model to large-scale video datasets \citep{tran_2015_c3d}. These spatiotemporal features are then fed to fully connected layers for generating the anomaly score. The backpropagation of the error is guided by the principle of multiple instance learning, allowing the model to learn anomalous segments despite having weak labels. This idea was taken up by \cite{zhu_2019_motion_aware} where, instead of using C3D, they made use of computing for the optical flows which are then fed to a temporal augmented network. Their proposed approach also makes use of an attention mechanism \citep{vaswani_2017_attention} that allows the network to identify which features are important to look at. Similarly, \cite{lin_2019_social_mil} also built upon this idea where they proposed a dual-branch network that incorporates motion into the initial network introduced by \cite{sultani_2018_real_world_anomaly}. The approach of \cite{lin_2019_social_mil} adapts the same network of \cite{sultani_2018_real_world_anomaly} as the first branch with a modification wherein an attention module \citep{vaswani_2017_attention} was added after the feature extraction layer. The second branch is similar in structure as the first branch except that it takes as an input social force maps \citep{mehran_2009_social_force} computed from the raw images to represent motion. 

Meanwhile, the approach of \cite{sabzalian_2019_deep_and_sparse_features} makes full use of the effectiveness of traditional and deep learning features for anomaly detection. Their proposed approach starts by identifying the foreground of the video by using optical flows. Once the regions of interest have been identified, a pre-trained Convolutional Neural Network is used to extract features alongside computing for traditional features like histogram of gradients and histogram of optical flows. These three features are combined by making use of an iteratively weighted nonnegative matrix factorization method \citep{sabzalian_2019_deep_and_sparse_features}. Afterwards, the features are clustered and the discrimination of whether or not the sample is an anomaly will be done via a voting system.

Aside from framing the problem as a regression problem, \cite{landi_2019_anomaly_locality} proposed to make use of locality when computing for the anomaly score. The approach is similar to that of \cite{sultani_2018_real_world_anomaly} except that their approach extracts a tube from the video which in a way localizes and adjusts the level of granularity when extracting features. From their experiments, they have shown that locality or, more specifically, zoning in on one region where the anomalous event takes place actually helps the method to accurately compute anomaly scores. 

Sparse coding for anomaly detection is an approach that learns a dictionary which attempts to encodes all normal events \citep{lu_2013_sparse_coding}. By revisiting sparse coding, \cite{luo_2019_sparse_coding_dnn} proposed temporally-coherent sparse coding to model the coherence between neighboring events for normal frames. These temporal features are then combined with spatial features learn from pre-trained networks across different scales for a normality score. Note that the features extracted pass through a Stacked Recurrent Neural Network autoencoder to generate the final features for scoring.

Past works demonstrated the effectiveness of autoencoders and that normal samples can be associated with at least one Gaussian Mixture Model. Because of this, \cite{fan_2020_gmm_cvae} proposed an end-to-end neural network called the Gaussian Mixture Fully Convolutional Variational Autoencoder to model anomalies and to predict them. Their model is trained on image and dynamic flow patches wherein both of them are separately fed into different networks. This basically captures separate motion and appearance features. Afterwards, joint probabilities are used to detect both appearance and motion anomalies via a sample energy-based method.

\section{Existing Benchmark Datasets}
This section discusses in detail the publicly available datasets for the task of anomaly detection. There are a few papers which have created their own datasets but most of the works have tried to at least use one benchmark dataset in order to evaluate the performance of their proposed approaches with respect to previously published works. A summary presenting a high-level view of all of the different datasets included in this subsection can be seen in Table \ref{benchmark_datasets}. Note that the dataset links are added as footnotes for reference. 

\begin{table}[]
\small 
\centering
\hspace*{-0.7cm}
\caption{Overview of Benchmark Datasets}
\label{benchmark_datasets}
\hspace*{-1.3cm}
\begin{tabular}{|c|l|l|l|l|l|}
\hline
\rowcolor[HTML]{000000} 
{\color[HTML]{FFFFFF} \textbf{Dataset}} & \multicolumn{1}{c|}{\cellcolor[HTML]{000000}{\color[HTML]{FFFFFF} \textbf{Frames}}} & \multicolumn{1}{c|}{\cellcolor[HTML]{000000}{\color[HTML]{FFFFFF} \textbf{Scene}}} & \multicolumn{1}{c|}{\cellcolor[HTML]{000000}{\color[HTML]{FFFFFF} \textbf{Labels}}} & \multicolumn{1}{c|}{\cellcolor[HTML]{000000}{\color[HTML]{FFFFFF} \textbf{Resolution}}} & \multicolumn{1}{c|}{\cellcolor[HTML]{000000}{\color[HTML]{FFFFFF} \textbf{Anomalies}}} \\ \hline
\begin{tabular}[c]{@{}c@{}}UCSD \\ Ped1\end{tabular} & 14,000 & Single & \begin{tabular}[c]{@{}l@{}}Spatial \&\\ Temporal\end{tabular} & 238$\times$158 & biker, cart, etc \\ \hline
\begin{tabular}[c]{@{}c@{}}UCSD \\ Ped2\end{tabular} & 4,560 & Single & \begin{tabular}[c]{@{}l@{}}Spatial \&\\ Temporal\end{tabular} & 360$\times$240 & biker, cart, etc \\ \hline
\begin{tabular}[c]{@{}c@{}}UMN \\ Lawn\end{tabular} & 1,450 & Single & Temporal & 320$\times$240 & escape panic \\ \hline
\begin{tabular}[c]{@{}c@{}}UMN \\ Indoor\end{tabular} & 4,415 & Single & Temporal & 320$\times$240 & escape panic \\ \hline
\begin{tabular}[c]{@{}c@{}}UMN \\ Plaza\end{tabular} & 2,145 & Single & Temporal & 320$\times$240 & escape panic \\ \hline
\begin{tabular}[c]{@{}c@{}}CUHK \\ Avenue\end{tabular} & 30,652 & Single & \begin{tabular}[c]{@{}l@{}}Spatial \&\\ Temporal\end{tabular} & 640$\times$360 & \begin{tabular}[c]{@{}l@{}}loitering, running,\\ throwing objects\end{tabular} \\ \hline
\begin{tabular}[c]{@{}c@{}}Subway \\ Entrance\end{tabular} & 72,401 & Single & Temporal & 512$\times$384 & \begin{tabular}[c]{@{}l@{}}avoiding payment, \\ wrong direction\end{tabular} \\ \hline
\begin{tabular}[c]{@{}c@{}}Subway\\ Exit\end{tabular} & 136,524 & Single & Temporal & 512$\times$384 & \begin{tabular}[c]{@{}l@{}}avoiding payment,\\ wrong direction\end{tabular} \\ \hline
\begin{tabular}[c]{@{}c@{}}Shanghai \\ Tech\end{tabular} & 317,398 & Multi & \begin{tabular}[c]{@{}l@{}}Spatial \&\\ Temporal\end{tabular} & 856$\times$480 & \begin{tabular}[c]{@{}l@{}}chasing, brawling\\ sudden motion, etc\end{tabular} \\ \hline
UCF-Crime & $\sim$13.8M & Multi & \begin{tabular}[c]{@{}l@{}}Video-level\\ \& Temporal\end{tabular} & 320$\times$240 & \begin{tabular}[c]{@{}l@{}}assault, burglary, \\ robbery, etc\end{tabular} \\ \hline
\begin{tabular}[c]{@{}c@{}}Street\\ Scene\end{tabular} & 203,257 & Single & \begin{tabular}[c]{@{}l@{}}Spatial \&\\ Temporal\end{tabular} & 1280$\times$720 & \begin{tabular}[c]{@{}l@{}}jaywalking, \\ person exits car, etc\end{tabular} \\ \hline
\end{tabular}
\end{table}

\subsection{The UCSD Pedestrian Dataset}

\begin{figure}[h]\centering
    \subfloat{\includegraphics[scale=0.4]{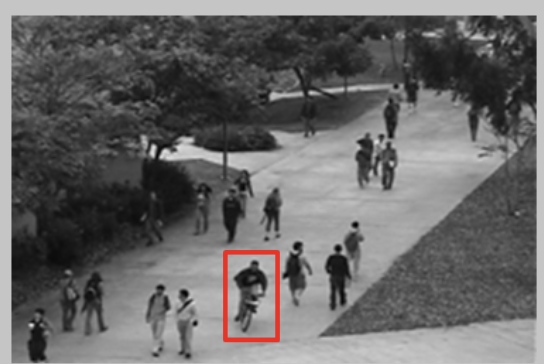}} 
    \subfloat{\includegraphics[scale=0.4]{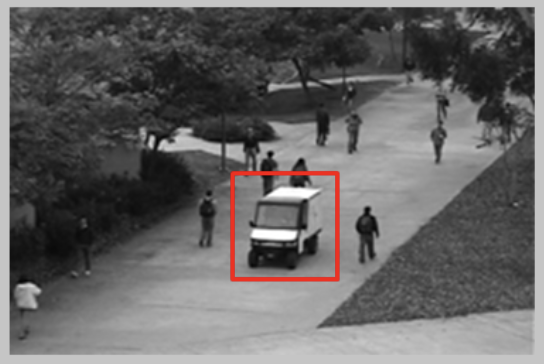}}
    \caption{USCD Pedestrian Example Anomalies}
    \label{example_ucsd}
\end{figure}

The UCSD Pedestrian dataset\footnote{http://www.svcl.ucsd.edu/projects/anomaly/dataset.html} was created by \cite{mahadevan_2010_ucsd} for the purpose of evaluating their approach on anomaly detection. The dataset contains videos overlooking pedestrian walkways taken by a stationary camera at 10 frames per second that is mounted at an elevation. In this dataset, anomalous events are either due to non-pedestrian entities in walkways or anomalous pedestrian motion. Some anomalous examples include bikers, skaters, cats, and the like. The dataset has two (2) subsets where each subset corresponds to a particular scene. The first scene includes people walking to and from the camera's angle while the second has people walking parallel to the camera plane. An example of the anomalies can be seen in Fig. \ref{example_ucsd}

The first subset called Peds1 contains 34 training clips and 36 testing clips having a resolution of 234 $\times$ 159. Meanwhile, the second subset called Peds2 contains 16 clips for training and 14 clips for testing having a resolution of 360 $\times$ 240. In general, there are around 3,400 frames with anomalies present while the normal frames are around 5,500. Both subsets have a frame-level ground truth and a pixel-level ground truth.

\subsection{The UMN Dataset}

The UMN dataset\footnote{http://mha.cs.umn.edu} has a total of 11 clips containing three (3) different scenes, specifically, a lawn scene, and indoor scene, and a plaza scene \citep{hu_2016_deep_incremental_sfa}. These video clips were captured at 30 frames per second using a stationary camera that has no significant illumination changes. The resolution of the captured video clips is at 320 $\times$ 240. With respect to the number of frames, all in all there are 7,740 frames where 1,450, 4,415, and 2,145 belong to lawn, indoor, and plaza scenes, respectively.

\begin{figure}[h]\centering
    \subfloat{\includegraphics[scale=0.33]{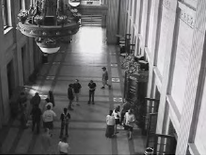}}
    \subfloat{\includegraphics[scale=0.33]{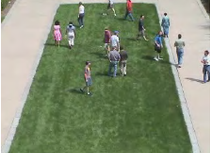}}
    \subfloat{\includegraphics[scale=0.33]{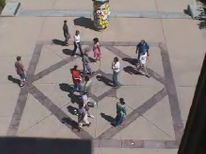}}
    \caption{UMN Dataset Examples}
    \label{example_umn}
\end{figure}

In this dataset, the particular anomaly that happens is when the people run to escape or when they panic. The sequences generally start with normal behavior where an escape panic behavior ensues. Sample frames from the dataset are shown in Fig. \ref{example_umn}.

\subsection{The CUHK Avenue Dataset}

Along with their proposed approach, \cite{lu_2013_sparse_coding} also created a dataset called the CUHK Avenue dataset\footnote{http://www.cse.cuhk.edu.hk/leojia/projects/detectabnormal/dataset.html} containing 16 videos for training and 21 videos for testing which includes 15,328 training frames and 15,324 testing frames with a resolution of 640 $\times$ 360. Furthermore, the dataset contains 47 different anomalies which include loitering, running, and throwing objects.

\begin{figure}[h]\centering
    \subfloat{\includegraphics[scale=0.3]{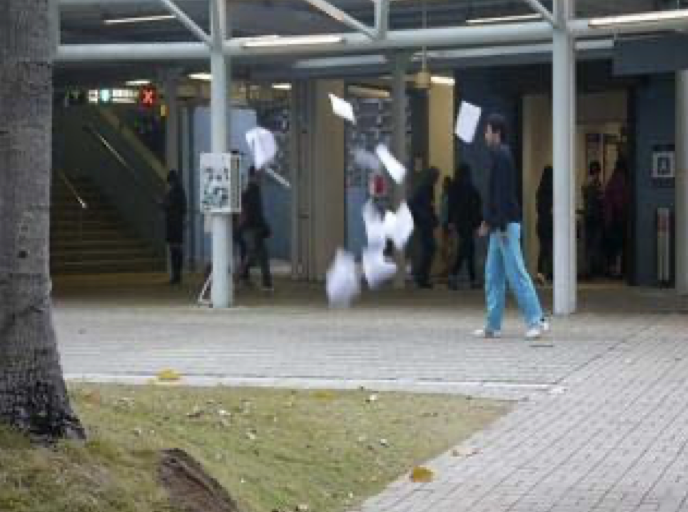}}
    \subfloat{\includegraphics[scale=0.3]{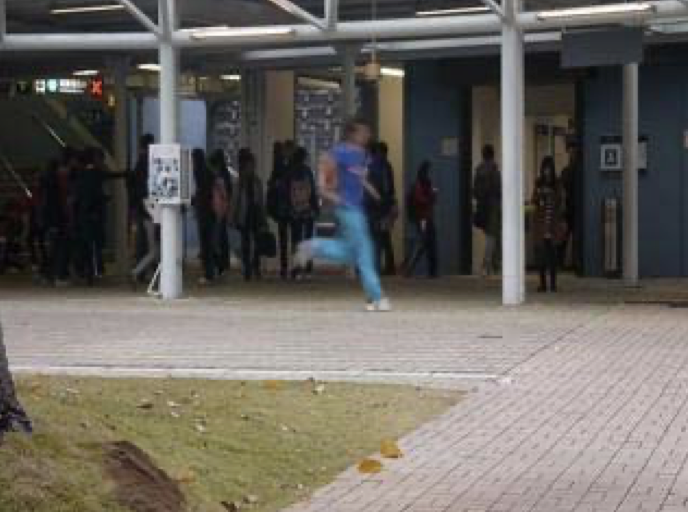}}
    \caption{CUHK Avenue Dataset}
    \label{example_cuhk}
\end{figure}

However, compared to the other datasets which have stationary cameras, the avenue dataset may have differences in camera angle and position. In addition, each of the videos is around 1 to 2 minutes long. Some example anomalies are shown in Fig. \ref{example_cuhk} where there is a running man on the left-hand side of the figure and the other image contains an anomalous action where paper is scattered around the area.

\subsection{The Subway Dataset}

The Subway Dataset\footnote{http://vision.eecs.yorku.ca/research/anomalous-behaviour-data/. This link only contains the Subway Exit} \citep{adam_2008_subway} contains two types of videos namely the "exit gate" and "entrace gate" videos. All in all, the videos are around two (2) hours long with a resolution of 512 $\times$ 384. 

\begin{figure}[h]\centering
    \subfloat{\includegraphics[scale=0.3]{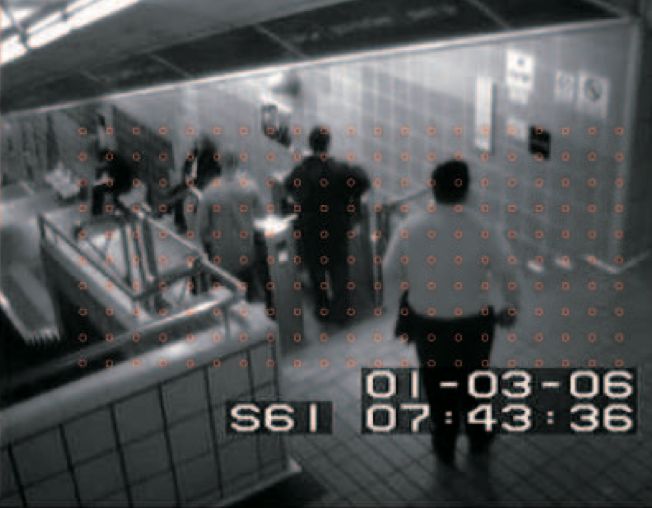}}
    \subfloat{\includegraphics[scale=0.3]{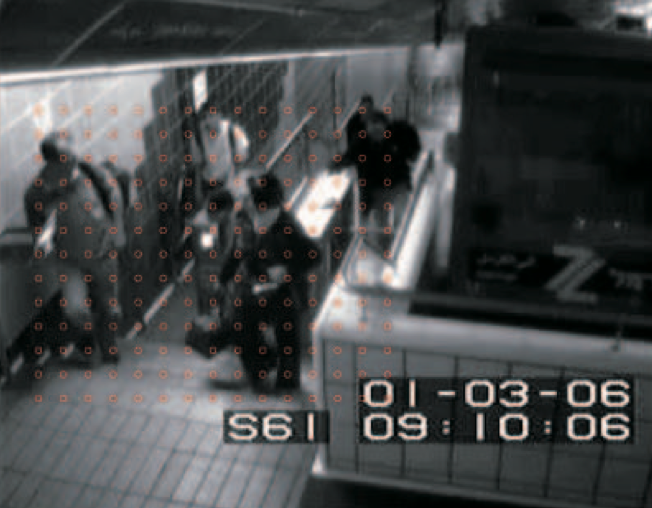}}
    \caption{Subway Dataset}
    \label{example_subway}
\end{figure}

The exit gate video has 136,524 frames while the entrance gate video has 72,401 frames \citep{liu_2018_future_frame}. In both scenarios, abnormality may include avoiding payment or walking in the wrong direction as the crowd. Comparing it to other datasets, the anomalies present in this dataset are relatively low \citep{sabokrou_2018_deep_anomaly}.

\subsection{The ShanghaiTech Campus Dataset}

The ShanghaiTech Campus\footnote{https://svip-lab.github.io/dataset/campus\_dataset.html} dataset \citep{liu_2018_future_frame} was proposed due to the lack of scene diversity from pre-existing benchmark datasets. Compared to previous datasets, the ShanghaiTech dataset has a larger number of videos having 330 training videos and 107 testing videos which consists of 13 different scenes and a large amount of varying anomaly types. The resolution of the videos in this dataset is at 856 $\times$ 480.

\begin{figure}[h]\centering
    \subfloat{\includegraphics[scale=0.2]{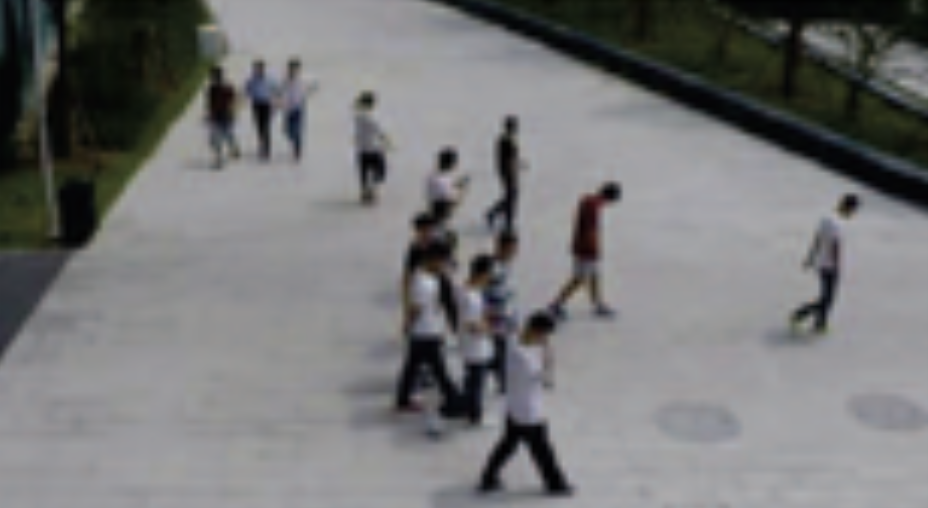}}
    \subfloat{\includegraphics[scale=0.2]{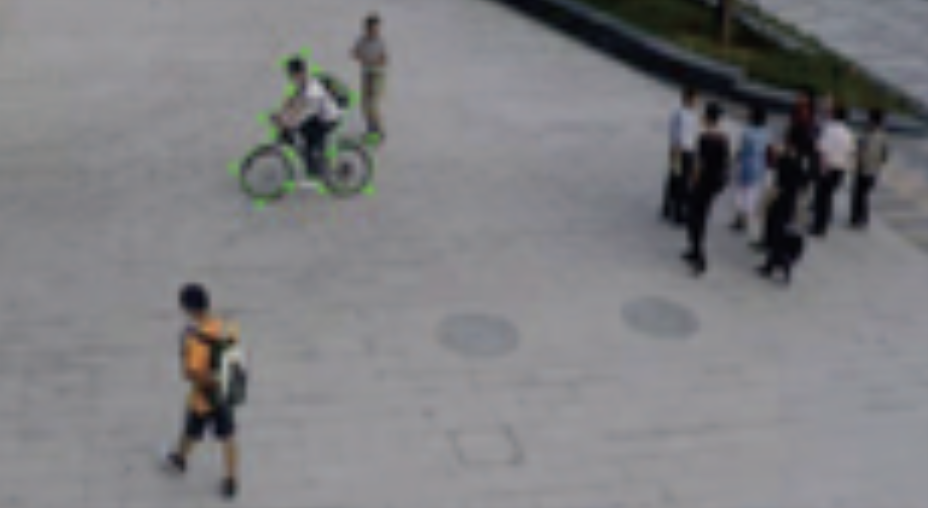}}
    \caption{ShanghaiTech Campus}
    \label{example_shanghai}
\end{figure}

An example is shown in Fig. \ref{example_shanghai} where the left image is the normal image with students walking while the right image contains the anomaly where there is a biker. Furthermore, there are also anomalies which are cause by sudden motion such as chasing and brawling. These types of anomalies are not included in datasets such was UCSD Pedestrian, CUHK Avenue, UMN Dataset, and Subway Dataset.

\subsection{The UCF-Crime Dataset}

Due to the previous datasets being relatively small in size, the UCF-Crime Dataset\footnote{https://webpages.uncc.edu/cchen62/dataset.html} was created by \cite{sultani_2018_real_world_anomaly}. This dataset contains 13 real-world anomalies namely accidents, burglary, explosion, fighting, robbery, shooting, stealing, shoplifting, and vandalism. Compared with previous datasets which were manually collected, this dataset was taken from Youtube\footnote{www.youtube.com} and LiveLeak\footnote{www.liveleak.com} using relevant text queries. These text queries are not limited to English, other languages (using Google Translate) were also used for searching. Overall, there are 950 untrimmed real-world surveillance videos and 950 normal videos garnering a total of 1,900 videos in the dataset. Note that the entire dataset has around 128 hours worth of data having a resolution of 240 $\times$ 320.

\begin{figure}[h]\centering
    \subfloat{\includegraphics[scale=0.4]{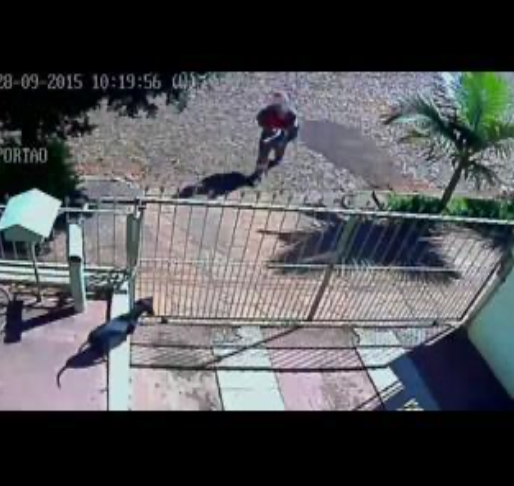}}
    \subfloat{\includegraphics[scale=0.4]{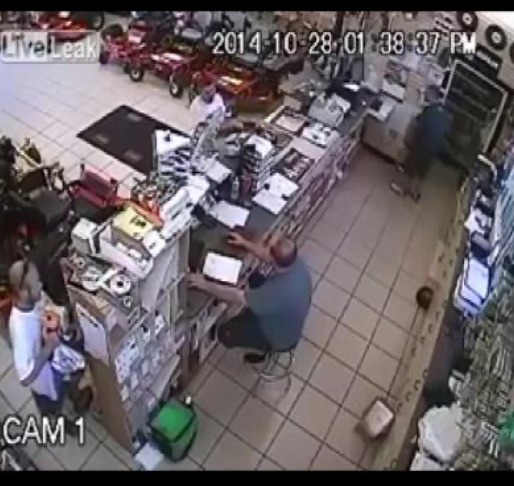}}
    \caption{UCF-Crime Dataset}
    \label{example_ucf}
\end{figure}

The dataset is already divided into training and test sets for uniformity. The training set consists of 810 anomalous videos while having 800 normal videos while the testing set has 150 normal and 140 anomalous videos. Despite being split into different datasets, all 13 anomalies are present in both sets lying at various locations in the video.

\subsection{The Street Scene Dataset}
One of the recently published datasets, the Street Scene dataset\footnote{http://www.merl.com/demos/video-anomaly-detection} \citep{ramachandra_2020_streetscene} was created to solve the existing problems that the older datasets were facing which is to have more realistic anomalies and to have a greater variety with respect to the types of anomalies that are present. In Street Scene, there are 46 training video sequences and 35 testing sequences. These videos are taken from a stationary USB camera which views a two-lane street that has pedestrian sidewalks and bike lanes.

\begin{figure}[h]\centering
    \subfloat{\includegraphics[scale=0.3]{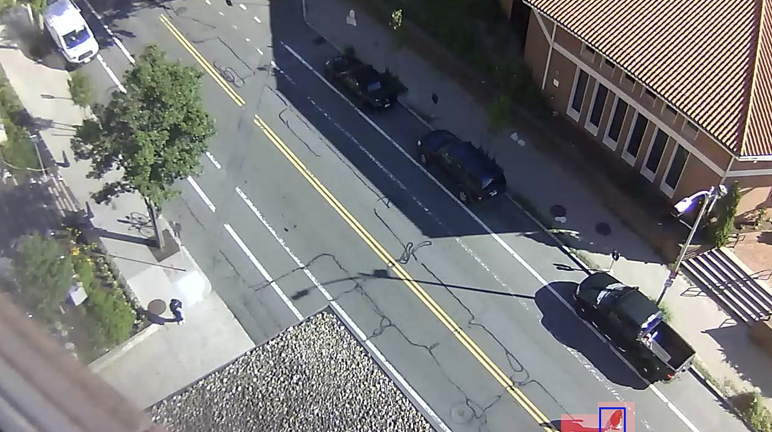}}
    \subfloat{\includegraphics[scale=0.3]{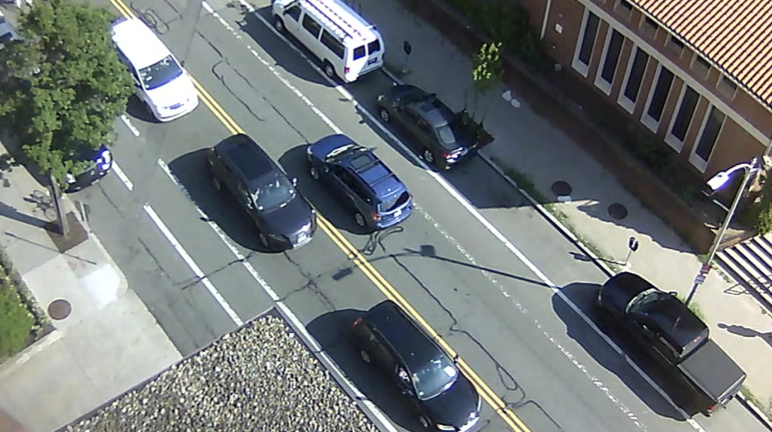}}
    \caption{Street Scene}
    \label{example_street}
\end{figure}

Example normal and anomaly in the Street Scene dataset are shown in Fig. \ref{example_street}. The left-hand side of the figure shows a person jaywalking which is an anommaly in the dataset while the right figure shows a normal scene. There are a total of 17 different anomaly types in the dataset namely jaywalking, biker outside lane, loitering, dog on sidewalk, car outside lane, biker on sidewalk, pedestrian reverses direction, and so on. 

\section{Evaluation Metrics}

This section briefly discusses the mostly used evaluation metrics by the papers that have been presented in this paper. Most of the works have followed the metrics introduced by \cite{li_2014_detection_in_crowded_scenes} where there are two (2) different criteria. The first one is a \textit{frame-level} criterion where a frame is considered anomalous if \textit{at least one of its pixels} are tagged as anomalous. To evaluate using the frame-level criterion, the temporal labels are used to determine metrics true positives and false positives. The second one is a \textit{pixel-level} criterion where if at least 40\% of the anomalous pixels are detected, the frame is considered to be anomalous. For both criterion, the area under the curve (AUC) of the receiver operating characteristic curve (ROC) is computed to measure the final performance of the models. Given a classification model having different thresholds, the receiver operating characteristic curve (ROC) illustrates the performance of the model. The true positive rate and false positive rates defined in Equations \ref{eq_tpr} and \ref{eq_fpr} are the parameters of the said curve \citep{bradley_1997_roc}. 

\begin{equation}\label{eq_tpr}
    \mbox{True Positive Rate} = \frac{\mbox{True Positives}}{\mbox{True Positives} + \mbox{False Negatives}}
\end{equation}
\begin{equation}\label{eq_fpr}
    \mbox{False Positive Rate} = \frac{\mbox{False Positives}}{\mbox{False Positives} + \mbox{True Negatives}}
\end{equation}

Basically, the ROC is a plot such that the true positive rate is on the y-axis and the false positive rate is on the x-axis. The values for each point in the plot is taken from different classification thresholds. The area under curve (AUC) of the ROC is used as a measure to determine how the good the model is performing. A higher value for the AUC of the ROC signifies that the model is performing well. The strengths of this metric include threshold-invariance and scale-invariance. It is scale-invariant because it does not look at the absolute values of the predictions and looks at how well the predictions are ranked. Meanwhile, it is also threshold-invariant since it measures the performance without considering the threshold chosen for classification. However, its strengths are also its weaknesses such as the scale-invariance of the metric might not be suited if well-calibrated probabilities are desired. Moreover, it is not suited for optimizing on metrics such as false positives in specific use cases since it expresses them as an aggregated value. Additionally, an equal error rate (EER) is computed alongside the receiver operating characteristic curve. The equal error rate computes for the percentage of misclassified frames when the false positive rate is equal to the miss rate. More specifically, it is when the $\mbox{False Positive Rate} = 1 - \mbox{True Positive Rate}$ for the frame-level criterion while it is $1 - \mbox{EER}$ for the pixel-level criterion \citep{li_2014_detection_in_crowded_scenes}.

There are problems in both of these metrics as mentioned in the work of \cite{ramachandra_2020_streetscene}. They have pointed out that in the frame-level criterion, an algorithm could still be considered correct even if the anomalous pixel doesn't necessarily overlap with the spatial region as to where the event is happening. Additionally, the pixel-level criterion does not take into account predictions that do not overlap with the ground truth. This prompted \cite{ramachandra_2020_streetscene} to propose new evaluation metrics alongside their recently published dataset. They have proposed to use track-based detection criterion and region-based detection criterion which they claim is similar to object tracking and object detection metrics. The track-based detection criterion measures the false positive regions per frame against the track-based detection rate (TBDR) which is defined in Equations \ref{eq_tbdr} and \ref{eq_fpr2}.

\begin{equation}\label{eq_tbdr}
    \mbox{TBDR} = \frac{\mbox{number of anomalous tracks detected}}{\mbox{total number of anomalous tracks}}
\end{equation}
\begin{equation}\label{eq_fpr2}
    \mbox{FPR} = \frac{\mbox{total false positive regions}}{\mbox{total frames}}
\end{equation}

Meanwhile, the region-based detection criterion measures the false positive regions per frame against the region-based detection rate (RBDR) across all testing frames. Correctly detected anomalous regions in frames are identified similar to the track-based detection criterion. The definition of RBDR is shown in Equation \ref{eq_rbdr}

\begin{equation}\label{eq_rbdr}
\mbox{RBDR} = \frac{\mbox{number of anomalous regions detected}}{\mbox{total number of anomalous regions}}
\end{equation}

Note that anomalous tracks are correctly identified if the ground truth has an intersection over union (IoU) above a threshold $\alpha$ with the detections. Similarly, anomalous regions in the frame is considered correctly identified if the ground truth has an IoU of above a threshold $\beta$ with the corresponding detected regions.

\section{Discussion}
Based on the different methodologies discussed in this paper, it is evident that anomaly detection is indeed a hard task. Several deep learning methods ranging from simple architectures to complex unified approaches have been proposed by different researchers. By categorizing the different approaches together into groups such as reconstruction error, future frame prediction, using classifiers, and scoring, a paradigm has been introduced on how to view anomaly detection approaches. Moreover, the variety of the type of approaches present also goes to show that researchers have been exploring different ways and thinking out of the box to determine anomalous events mainly because of its difficulty.

One common theme from all of the papers is that most of them still are careful about taking into account several aspects of human action such as appearance and motion. Representations may differ such as the work of \cite{lin_2019_social_mil} which uses social force maps while \cite{xu_2020_adaptive_intraframe} uses optical flows but the main idea remains the same. This points the research community to a direction that appearance and motion play a big part in detecting anomalies. More so, that even in deep learning approaches (which is supposed to automatically learn discriminative features), researchers still make use of these features or concepts to guide the network and make it look properly at these specific variables.

Recent papers have started to think of creating end-to-end deep learning solutions and unified architectures rather than making use of separate components in a traditional pipeline. This is important as well because end-to-end deep learning solutions are easily deployable in real-life, making the research more accessible and more usable than it is now. However, end-to-end deep learning solutions require a lot of data which might be a problem for older datasets such as the UCSD Pedestrian or UMN Dataset but large scale datasets have been proposed by \cite{sultani_2018_real_world_anomaly, ramachandra_2020_streetscene, liu_2018_future_frame} to help solve this problem. Yet, an important issue to also consider as well is that video data is very laborious to annotate and collect which one of the main reasons why there haven't been as much large scale datasets published yet despite having tons of data publicly available in video sharing sites. This stresses the importance of making use of unsupervised or weakly-supervised approaches in tackling this problem.

With regards to evaluation, as presented by \cite{ramachandra_2020_streetscene}, the current evaluation metrics using the frame-level criterion and pixel-level criterion might not be representative of the performance of the model due to the reasons stated in their work. Hence, there might be a need to have more robust evaluation metrics which would be more effective irrespective of the type of new datasets that might be published in the future. Future evaluation metrics must consider providing better ways to assess spatial aspects of future methodologies since it is important to know which part of the frames cause the anomalies. This in turn, allows faster and better inference to what is happening should the approaches be deployed in real life.

Looking from a different perspective, results have become better over time because methods by various researchers, have successfully managed to incorporate spatial and temporal information to their models, thereby achieving excellent results. Yet, for real-life anomalous events, it is more than spatial and temporal information, there also needs to be context added to make the models more robust. As seen from the different definitions of different authors, the very definition of what an anomaly is also vary from one context to another. One possible way to achieve this is to slowly pivot the research area towards larger datasets and datasets captured from real-life videos and real-life scenarios. Furthermore, borrowing concepts such as attention or transformers from different fields might also be helpful to achieve this goal.

\section{Conclusions}

This paper has provided an overview of the recent advances in anomaly detection for videos specifically using deep learning techniques. Four types of categories of current approaches have been introduced with respect to the final step in identifying anomalies such as using reconstruction error, predicting future frames, using classification, or using scoring. These categories show the diversity of the approaches and it also is a testament to the difficulty of the problem as it forces researchers and practitioners alike to think out of the box to find better solutions to the problem. 

In addition, this paper has also presented the different commonly used datasets along with important details such as the video resolution and example anomalies found within the respective datasets. Over time, it can be seen that the datasets are gradually increasing in size and are also becoming closer to real-life scenarios. However, there is still an issue of manually annotating these videos. Approaches that leverage weakly-supervised or unsupervised learning should be explored more in the hopes that it might also be able to automatically annotate videos once they learn from a small sample.

Future areas of research might include adding context since most of the works have been successful in modelling both motion and appearance, studying the recently published large-scale datasets, creating end-to-end deep learning frameworks, and focusing more on approaches that require little to no supervision.


%
%


\bibliographystyle{spbasic}
\bibliography{references}

\begin{thebibliography}{58}
\providecommand{\natexlab}[1]{#1}
\providecommand{\url}[1]{{#1}}
\providecommand{\urlprefix}{URL }
\expandafter\ifx\csname urlstyle\endcsname\relax
  \providecommand{\doi}[1]{DOI~\discretionary{}{}{}#1}\else
  \providecommand{\doi}{DOI~\discretionary{}{}{}\begingroup
  \urlstyle{rm}\Url}\fi
\providecommand{\eprint}[2][]{\url{#2}}

\bibitem[{{Adam} et~al.(2008){Adam}, {Rivlin}, {Shimshoni}, and
  {Reinitz}}]{adam_2008_subway}
{Adam} A, {Rivlin} E, {Shimshoni} I, {Reinitz} D (2008) Robust real-time
  unusual event detection using multiple fixed-location monitors. IEEE
  Transactions on Pattern Analysis and Machine Intelligence 30(3):555--560

\bibitem[{Baldi(2011)}]{baldi_2011_autoencoders}
Baldi P (2011) Autoencoders, unsupervised learning and deep architectures. In:
  Proceedings of the 2011 International Conference on Unsupervised and Transfer
  Learning Workshop - Volume 27, JMLR.org, UTLW’11, p 37–50

\bibitem[{Benezeth et~al.(2009)Benezeth, Jodoin, Saligrama, and
  Rosenberger}]{benezeth_2009_spatiotemporal_cooccurrences}
Benezeth Y, Jodoin PM, Saligrama V, Rosenberger C (2009) Abnormal events
  detection based on spatio-temporal co-occurences. In: 2009 {IEEE} Conference
  on Computer Vision and Pattern Recognition, {IEEE},
  \doi{10.1109/cvpr.2009.5206686},
  \urlprefix\url{https://doi.org/10.1109/cvpr.2009.5206686}

\bibitem[{Bradley(1997)}]{bradley_1997_roc}
Bradley AP (1997) The use of the area under the roc curve in the evaluation of
  machine learning algorithms. Pattern Recognition 30(7):1145–1159,
  \doi{10.1016/s0031-3203(96)00142-2}

\bibitem[{Calderara et~al.(2011)Calderara, Heinemann, Prati, Cucchiara, and
  Tishby}]{calderara_2011_spectral_graph}
Calderara S, Heinemann U, Prati A, Cucchiara R, Tishby N (2011) Detecting
  anomalies in people's trajectories using spectral graph analysis.
  \urlprefix\url{https://www.sciencedirect.com/science/article/pii/S1077314211000919}

\bibitem[{Chandola et~al.(2009)Chandola, Banerjee, and
  Kumar}]{chandola_2019_anomaly_detection}
Chandola V, Banerjee A, Kumar V (2009) Anomaly detection: A survey. ACM Comput
  Surv 41(3), \doi{10.1145/1541880.1541882},
  \urlprefix\url{https://doi.org/10.1145/1541880.1541882}

\bibitem[{Fan et~al.(2020)Fan, Wen, Li, Qiu, Levine, and
  Xiao}]{fan_2020_gmm_cvae}
Fan Y, Wen G, Li D, Qiu S, Levine MD, Xiao F (2020) Video anomaly detection and
  localization via gaussian mixture fully convolutional variational
  autoencoder. Computer Vision and Image Understanding 195:102920,
  \doi{https://doi.org/10.1016/j.cviu.2020.102920},
  \urlprefix\url{http://www.sciencedirect.com/science/article/pii/S1077314218302674}

\bibitem[{Gong et~al.(2019)Gong, Liu, Le, Saha, Mansour, Venkatesh, and
  Hengel}]{gong_2019_memory_deep_ae}
Gong D, Liu L, Le V, Saha B, Mansour M, Venkatesh S, Hengel A (2019) Memorizing
  normality to detect anomaly: Memory-augmented deep autoencoder for
  unsupervised anomaly detection. 2019 IEEE/CVF Conference on Computer Vision
  and Pattern Recognition pp 1705--1714, \doi{10.1109/ICCV.2019.00179}

\bibitem[{Goodfellow et~al.(2014)Goodfellow, Pouget-Abadie, Mirza, Xu,
  Warde-Farley, Ozair, Courville, and Bengio}]{goodfellow_2014_gan}
Goodfellow IJ, Pouget-Abadie J, Mirza M, Xu B, Warde-Farley D, Ozair S,
  Courville A, Bengio Y (2014) Generative adversarial nets. In: Proceedings of
  the 27th International Conference on Neural Information Processing Systems -
  Volume 2, MIT Press, Cambridge, MA, USA, NIPS’14, p 2672–2680

\bibitem[{Hasan et~al.(2016)Hasan, Choi, Neumann, Roy-Chowdhury, and
  Davis}]{hasan_2016_temporal_regularity}
Hasan M, Choi J, Neumann J, Roy-Chowdhury AK, Davis LS (2016) Learning temporal
  regularity in video sequences. 2016 IEEE Conference on Computer Vision and
  Pattern Recognition (CVPR) pp 733--742

\bibitem[{Hochreiter and Schmidhuber(1997)}]{hochreiter_1997_lstm}
Hochreiter S, Schmidhuber J (1997) Long short-term memory. Neural Comput
  9(8):1735–1780, \doi{10.1162/neco.1997.9.8.1735},
  \urlprefix\url{https://doi.org/10.1162/neco.1997.9.8.1735}

\bibitem[{{Hu} et~al.(2016){Hu}, {Hu}, {Huang}, {Zhang}, and
  {Wu}}]{hu_2016_deep_incremental_sfa}
{Hu} X, {Hu} S, {Huang} Y, {Zhang} H, {Wu} H (2016) Video anomaly detection
  using deep incremental slow feature analysis network. IET Computer Vision
  10(4):258--265

\bibitem[{Ionescu et~al.(2019)Ionescu, Khan, Georgescu, and
  Shao}]{ionescu_2019_object_centric}
Ionescu RT, Khan F, Georgescu M, Shao L (2019) Object-centric auto-encoders and
  dummy anomalies for abnormal event detection in video. 2019 IEEE/CVF
  Conference on Computer Vision and Pattern Recognition (CVPR) pp 7834--7843,
  \doi{10.1109/CVPR.2019.00803}

\bibitem[{Jiang et~al.(2011)Jiang, Yuan, Tsaftaris, and
  Katsaggelos}]{fan_2011_spatiotemporal_context}
Jiang F, Yuan J, Tsaftaris SA, Katsaggelos AK (2011) Anomalous video event
  detection using spatiotemporal context.
  \urlprefix\url{https://dl.acm.org/doi/10.1016/j.cviu.2010.10.008}

\bibitem[{Kaur et~al.(2018)Kaur, Gangadharappa, and
  Gautam}]{kaur_2018_overview_of_anomaly_detection}
Kaur P, Gangadharappa M, Gautam S (2018) An overview of anomaly detection in
  video surveillance. In: 2018 International Conference on Advances in
  Computing, Communication Control and Networking ({ICACCCN}), {IEEE},
  \doi{10.1109/icacccn.2018.8748454},
  \urlprefix\url{https://doi.org/10.1109/icacccn.2018.8748454}

\bibitem[{Kim and Grauman(2009)}]{kim_2009_spacetime_mrf}
Kim J, Grauman K (2009) Observe locally, infer globally: A space-time {MRF} for
  detecting abnormal activities with incremental updates. In: 2009 {IEEE}
  Conference on Computer Vision and Pattern Recognition, {IEEE},
  \doi{10.1109/cvpr.2009.5206569},
  \urlprefix\url{https://doi.org/10.1109/cvpr.2009.5206569}

\bibitem[{Krizhevsky et~al.(2017)Krizhevsky, Sutskever, and
  Hinton}]{krizhevsky_2012_imagenet_classification}
Krizhevsky A, Sutskever I, Hinton GE (2017) Imagenet classification with deep
  convolutional neural networks. Commun ACM 60(6):84–90,
  \doi{10.1145/3065386}, \urlprefix\url{https://doi.org/10.1145/3065386}

\bibitem[{Landi et~al.(2019)Landi, Snoek, and
  Cucchiara}]{landi_2019_anomaly_locality}
Landi F, Snoek CGM, Cucchiara R (2019) Anomaly locality in video surveillance.
  ArXiv abs/1901.10364

\bibitem[{Li et~al.(2013)Li, Han, Ye, and
  Jiao}]{li_2013_trajectory_sparse_reconstruction}
Li C, Han Z, Ye Q, Jiao J (2013) Visual abnormal behavior detection based on
  trajectory sparse reconstruction analysis.
  \urlprefix\url{https://www.sciencedirect.com/science/article/abs/pii/S0925231213000179}

\bibitem[{Li et~al.(2014)Li, Mahadevan, and
  Vasconcelos}]{li_2014_detection_in_crowded_scenes}
Li W, Mahadevan V, Vasconcelos N (2014) Anomaly detection and localization in
  crowded scenes. {IEEE} Transactions on Pattern Analysis and Machine
  Intelligence 36(1):18--32, \doi{10.1109/tpami.2013.111},
  \urlprefix\url{https://doi.org/10.1109/tpami.2013.111}

\bibitem[{Li and min Cai(2016)}]{li_2016_anomaly_detection_techniques}
Li X, min Cai Z (2016) Anomaly detection techniques in surveillance videos. In:
  2016 9th International Congress on Image and Signal Processing, {BioMedical}
  Engineering and Informatics ({CISP}-{BMEI}), {IEEE},
  \doi{10.1109/cisp-bmei.2016.7852681},
  \urlprefix\url{https://doi.org/10.1109/cisp-bmei.2016.7852681}

\bibitem[{{Lin} et~al.(2019){Lin}, {Yang}, {Tang}, {Shi}, and
  {Chen}}]{lin_2019_social_mil}
{Lin} S, {Yang} H, {Tang} X, {Shi} T, {Chen} L (2019) Social mil:
  Interaction-aware for crowd anomaly detection. In: 2019 16th IEEE
  International Conference on Advanced Video and Signal Based Surveillance
  (AVSS), pp 1--8

\bibitem[{Liu et~al.(2016)Liu, Anguelov, Erhan, Szegedy, Reed, Fu, and
  Berg}]{liu_2016_ssd}
Liu W, Anguelov D, Erhan D, Szegedy C, Reed S, Fu CY, Berg AC (2016) {SSD}:
  Single shot {MultiBox} detector. In: Computer Vision {\textendash} {ECCV}
  2016, Springer International Publishing, pp 21--37,
  \doi{10.1007/978-3-319-46448-0\_2},
  \urlprefix\url{https://doi.org/10.1007/978-3-319-46448-0\_2}

\bibitem[{Liu et~al.(2018)Liu, Luo, Lian, and Gao}]{liu_2018_future_frame}
Liu W, Luo W, Lian D, Gao S (2018) Future frame prediction for anomaly
  detection - a new baseline. 2018 IEEE/CVF Conference on Computer Vision and
  Pattern Recognition pp 6536--6545

\bibitem[{Lopes et~al.(2017)Lopes, de~Aguiar, De~Souza, and
  Oliveira-Santos}]{lopes_2017_cnn_face}
Lopes AT, de~Aguiar E, De~Souza AF, Oliveira-Santos T (2017) Facial expression
  recognition with convolutional neural networks. Pattern Recogn
  61(C):610–628, \doi{10.1016/j.patcog.2016.07.026},
  \urlprefix\url{https://doi.org/10.1016/j.patcog.2016.07.026}

\bibitem[{{Lu} et~al.(2013){Lu}, {Shi}, and {Jia}}]{lu_2013_sparse_coding}
{Lu} C, {Shi} J, {Jia} J (2013) Abnormal event detection at 150 fps in matlab.
  In: 2013 IEEE International Conference on Computer Vision, pp 2720--2727

\bibitem[{{Luo} et~al.(2019){Luo}, {Liu}, {Lian}, {Tang}, {Duan}, {Peng}, and
  {Gao}}]{luo_2019_sparse_coding_dnn}
{Luo} W, {Liu} W, {Lian} D, {Tang} J, {Duan} L, {Peng} X, {Gao} S (2019) Video
  anomaly detection with sparse coding inspired deep neural networks. IEEE
  Transactions on Pattern Analysis and Machine Intelligence pp 1--1

\bibitem[{Mahadevan et~al.(2010)Mahadevan, LI, Bhalodia, and
  Vasconcelos}]{mahadevan_2010_ucsd}
Mahadevan V, LI WX, Bhalodia V, Vasconcelos N (2010) Anomaly detection in
  crowded scenes. In: Proceedings of IEEE Conference on Computer Vision and
  Pattern Recognition, pp 1975--1981

\bibitem[{Malinowski et~al.(2017)Malinowski, Rohrbach, and
  Fritz}]{malinowski_2017_vqa}
Malinowski M, Rohrbach M, Fritz M (2017) Ask your neurons: A deep learning
  approach to visual question answering. Int J Comput Vision
  125(1–3):110–135, \doi{10.1007/s11263-017-1038-2},
  \urlprefix\url{https://doi.org/10.1007/s11263-017-1038-2}

\bibitem[{Massi(2019)}]{massi_2019_pic_ae}
Massi M (2019) Wikimedia Commons.
  \urlprefix\url{https://commons.wikimedia.org/wiki/File:Autoencoder\_schema.png}

\bibitem[{Medel and Savakis(2016)}]{medel_2016_predictive_cnn_lstm}
Medel JR, Savakis AE (2016) Anomaly detection in video using predictive
  convolutional long short-term memory networks. CoRR abs/1612.00390,
  \urlprefix\url{http://arxiv.org/abs/1612.00390}, \eprint{1612.00390}

\bibitem[{Mehran et~al.(2009)Mehran, Oyama, and
  Shah}]{mehran_2009_social_force}
Mehran R, Oyama A, Shah M (2009) Abnormal crowd behavior detection using social
  force model. In: 2009 {IEEE} Conference on Computer Vision and Pattern
  Recognition, {IEEE}, \doi{10.1109/cvpr.2009.5206641},
  \urlprefix\url{https://doi.org/10.1109/cvpr.2009.5206641}

\bibitem[{Mei and Zhang(2017)}]{mei_2017_deep_learning_for_video}
Mei T, Zhang C (2017) Deep learning for intelligent video analysis. In:
  Proceedings of the 25th ACM International Conference on Multimedia,
  Association for Computing Machinery, New York, NY, USA, MM ’17, p
  1955–1956, \doi{10.1145/3123266.3130141},
  \urlprefix\url{https://doi.org/10.1145/3123266.3130141}

\bibitem[{Narasimhan and S.(2018)}]{narasimhan_2018_sparse_denoising_ae}
Narasimhan MG, S SK (2018) Dynamic video anomaly detection and localization
  using sparse denoising autoencoders. Multimedia Tools Appl
  77(11):13173–13195, \doi{10.1007/s11042-017-4940-2},
  \urlprefix\url{https://doi.org/10.1007/s11042-017-4940-2}

\bibitem[{{Pan} et~al.(2011){Pan}, {Tsang}, {Kwok}, and
  {Yang}}]{pan_2011_transfer_component_analysis}
{Pan} SJ, {Tsang} IW, {Kwok} JT, {Yang} Q (2011) Domain adaptation via transfer
  component analysis. IEEE Transactions on Neural Networks 22(2):199--210

\bibitem[{{Popoola} and {Wang}(2012)}]{popola_2012_abnormal_review}
{Popoola} OP, {Wang} K (2012) Video-based abnormal human behavior
  recognition—a review. IEEE Transactions on Systems, Man, and Cybernetics,
  Part C (Applications and Reviews) 42(6):865--878

\bibitem[{Ramachandra and Jones(2020)}]{ramachandra_2020_streetscene}
Ramachandra B, Jones MJ (2020) Street scene: A new dataset and evaluation
  protocol for video anomaly detection. In: 2020 {IEEE} Winter Conference on
  Applications of Computer Vision ({WACV}), {IEEE},
  \doi{10.1109/wacv45572.2020.9093457},
  \urlprefix\url{https://doi.org/10.1109/wacv45572.2020.9093457}

\bibitem[{Ribeiro et~al.(2018)Ribeiro, Lazzaretti, and
  Lopes}]{ribeiro_2018_deep_ae}
Ribeiro M, Lazzaretti AE, Lopes HS (2018) A study of deep convolutional
  auto-encoders for anomaly detection in videos. Pattern Recognition Letters
  105:13 -- 22, \doi{https://doi.org/10.1016/j.patrec.2017.07.016},
  \urlprefix\url{http://www.sciencedirect.com/science/article/pii/S0167865517302489},
  machine Learning and Applications in Artificial Intelligence

\bibitem[{Ronneberger et~al.(2015)Ronneberger, Fischer, and
  Brox}]{ronneberger_2015_unet}
Ronneberger O, Fischer P, Brox T (2015) U-net: Convolutional networks for
  biomedical image segmentation. In: Lecture Notes in Computer Science,
  Springer International Publishing, pp 234--241,
  \doi{10.1007/978-3-319-24574-4\_28},
  \urlprefix\url{https://doi.org/10.1007/978-3-319-24574-4\_28}

\bibitem[{Sabokrou et~al.(2016)Sabokrou, Fathy, and
  Hoseini}]{sabokrou_2016_sparsity_reconstruction_error_ae}
Sabokrou M, Fathy M, Hoseini M (2016) Video anomaly detection and localisation
  based on the sparsity and reconstruction error of auto-encoder. Electronics
  Letters 52:1122--1124

\bibitem[{{Sabokrou} et~al.(2017){Sabokrou}, {Fayyaz}, {Fathy}, and
  {Klette}}]{sabokrou_2017_deep_cascade}
{Sabokrou} M, {Fayyaz} M, {Fathy} M, {Klette} R (2017) Deep-cascade: Cascading
  3d deep neural networks for fast anomaly detection and localization in
  crowded scenes. IEEE Transactions on Image Processing 26(4):1992--2004

\bibitem[{Sabokrou et~al.(2018)Sabokrou, Fayyaz, Fathy, Moayed, and
  Klette}]{sabokrou_2018_deep_anomaly}
Sabokrou M, Fayyaz M, Fathy M, Moayed Z, Klette R (2018) Deep-anomaly: Fully
  convolutional neural network for fast anomaly detection in crowded scenes.
  Computer Vision and Image Understanding 172:88 -- 97,
  \doi{https://doi.org/10.1016/j.cviu.2018.02.006},
  \urlprefix\url{http://www.sciencedirect.com/science/article/pii/S1077314218300249}

\bibitem[{{Sabzalian} et~al.(2019){Sabzalian}, {Marvi}, and
  {Ahmadyfard}}]{sabzalian_2019_deep_and_sparse_features}
{Sabzalian} B, {Marvi} H, {Ahmadyfard} A (2019) Deep and sparse features for
  anomaly detection and localization in video. In: 2019 4th International
  Conference on Pattern Recognition and Image Analysis (IPRIA), pp 173--178

\bibitem[{[dos Santos] et~al.(2019)[dos Santos], Ribeiro, and
  Ponti}]{dossantos_2019_generalization_of_feature_embeddings}
[dos Santos] FP, Ribeiro LS, Ponti MA (2019) Generalization of feature
  embeddings transferred from different video anomaly detection domains.
  Journal of Visual Communication and Image Representation 60:407 -- 416,
  \doi{https://doi.org/10.1016/j.jvcir.2019.02.035},
  \urlprefix\url{http://www.sciencedirect.com/science/article/pii/S1047320319300926}

\bibitem[{{Sultani} et~al.(2018){Sultani}, {Chen}, and
  {Shah}}]{sultani_2018_real_world_anomaly}
{Sultani} W, {Chen} C, {Shah} M (2018) Real-world anomaly detection in
  surveillance videos. In: 2018 IEEE/CVF Conference on Computer Vision and
  Pattern Recognition, pp 6479--6488

\bibitem[{{Tran} et~al.(2015){Tran}, {Bourdev}, {Fergus}, {Torresani}, and
  {Paluri}}]{tran_2015_c3d}
{Tran} D, {Bourdev} L, {Fergus} R, {Torresani} L, {Paluri} M (2015) Learning
  spatiotemporal features with 3d convolutional networks. In: 2015 IEEE
  International Conference on Computer Vision (ICCV), pp 4489--4497

\bibitem[{Tung et~al.(2010)Tung, Zelek, and
  Clausi}]{tung_2010_goal_based_trajectory}
Tung F, Zelek JS, Clausi DA (2010) Goal-based trajectory analysis for unusual
  behaviour detection in intelligent surveillance.
  \urlprefix\url{https://www.sciencedirect.com/science/article/abs/pii/S026288561000154X}

\bibitem[{Vaswani et~al.(2017)Vaswani, Shazeer, Parmar, Uszkoreit, Jones,
  Gomez, Kaiser, and Polosukhin}]{vaswani_2017_attention}
Vaswani A, Shazeer N, Parmar N, Uszkoreit J, Jones L, Gomez AN, Kaiser u,
  Polosukhin I (2017) Attention is all you need. In: Proceedings of the 31st
  International Conference on Neural Information Processing Systems, Curran
  Associates Inc., Red Hook, NY, USA, NIPS’17, p 6000–6010

\bibitem[{Vu et~al.(2017)Vu, Phung, Nguyen, Trevors, and
  Venkatesh}]{vu_2017_energy_based_models}
Vu H, Phung D, Nguyen TD, Trevors A, Venkatesh S (2017) Energy-based models for
  video anomaly detection. \eprint{1708.05211}

\bibitem[{Wang et~al.(2018)Wang, Zhu, Yin, and
  Porikli}]{wang_2018_local_motion_ocelm}
Wang S, Zhu E, Yin J, Porikli F (2018) Video anomaly detection and localization
  by local motion based joint video representation and {OCELM}. Neurocomputing
  277:161--175, \doi{10.1016/j.neucom.2016.08.156},
  \urlprefix\url{https://doi.org/10.1016/j.neucom.2016.08.156}

\bibitem[{Wiskott and Sejnowski(2002)}]{wiskott_2002_slow_feature_analysis}
Wiskott L, Sejnowski TJ (2002) Slow feature analysis: Unsupervised learning of
  invariances. Neural Computation 14(4):715--770,
  \doi{10.1162/089976602317318938},
  \urlprefix\url{https://doi.org/10.1162/089976602317318938}

\bibitem[{{Xu} et~al.(2020){Xu}, {Sun}, and
  {Jiang}}]{xu_2020_adaptive_intraframe}
{Xu} K, {Sun} T, {Jiang} X (2020) Video anomaly detection and localization
  based on an adaptive intra-frame classification network. IEEE Transactions on
  Multimedia 22(2):394--406

\bibitem[{Yan et~al.(2016)Yan, Liu, and Hong}]{yan_2016_deep_cnn_feature}
Yan H, Liu X, Hong R (2016) Image classification via fusing the latent deep cnn
  feature. In: Proceedings of the International Conference on Internet
  Multimedia Computing and Service, Association for Computing Machinery, New
  York, NY, USA, ICIMCS’16, p 110–113, \doi{10.1145/3007669.3007706},
  \urlprefix\url{https://doi.org/10.1145/3007669.3007706}

\bibitem[{Zhang et~al.(2016)Zhang, Lu, Zhang, and
  Ruan}]{zhang_2016_combining_appearance_and_motion}
Zhang Y, Lu H, Zhang L, Ruan X (2016) Combining motion and appearance cues for
  anomaly detection. Pattern Recognition 51:443--452,
  \doi{10.1016/j.patcog.2015.09.005},
  \urlprefix\url{https://doi.org/10.1016/j.patcog.2015.09.005}

\bibitem[{Zhao et~al.(2017)Zhao, Deng, Shen, Liu, Lu, and
  Hua}]{zhao_2017_spatiotemporal_ae}
Zhao Y, Deng B, Shen C, Liu Y, Lu H, Hua XS (2017) Spatio-temporal autoencoder
  for video anomaly detection. In: Proceedings of the 25th ACM International
  Conference on Multimedia, Association for Computing Machinery, New York, NY,
  USA, MM ’17, p 1933–1941, \doi{10.1145/3123266.3123451},
  \urlprefix\url{https://doi.org/10.1145/3123266.3123451}

\bibitem[{{Zhou} et~al.(2019){Zhou}, {Du}, {Zhu}, {Peng}, {Liu}, and
  {Goh}}]{zhou_2019_anomalynet}
{Zhou} JT, {Du} J, {Zhu} H, {Peng} X, {Liu} Y, {Goh} RSM (2019) Anomalynet: An
  anomaly detection network for video surveillance. IEEE Transactions on
  Information Forensics and Security 14(10):2537--2550

\bibitem[{Zhu and Newsam(2019)}]{zhu_2019_motion_aware}
Zhu Y, Newsam S (2019) Motion-aware feature for improved video anomaly
  detection. British Machine Vision Conference

\bibitem[{Zoph et~al.(2018)Zoph, Vasudevan, Shlens, and
  Le}]{zoph_2018_transferrable_architectures}
Zoph B, Vasudevan V, Shlens J, Le QV (2018) Learning transferable architectures
  for scalable image recognition. In: 2018 {IEEE}/{CVF} Conference on Computer
  Vision and Pattern Recognition, {IEEE}, \doi{10.1109/cvpr.2018.00907},
  \urlprefix\url{https://doi.org/10.1109/cvpr.2018.00907}

\end{thebibliography}

\end{document}